\documentclass[runningheads]{llncs}
\usepackage[T1]{fontenc}
\usepackage{amsmath} 
\usepackage{graphicx}
\usepackage{booktabs}
\usepackage{multirow} 
\usepackage{float} 
\usepackage{graphicx}
\usepackage{soul}
\usepackage{hyperref}

\begin{document}
\title{AutoML-guided Fusion of Entity and \\LLM-based Representations for Document Classification}
%

%
\titlerunning{AutoML-guided Entity and LLM-based Representations}
%
\author{Boshko Koloski\inst{1,2}\orcidID{0000-0002-7330-0579} \and
Senja Pollak\inst{1}\orcidID{0000-0002-4380-0863} \and
Roberto Navigli \inst{3}\orcidID{0000-0003-3831-9706} \and
Bla\v{z} \v{S}krlj\inst{1}\orcidID{0000-0002-9916-8756} 
}
\authorrunning{B. Koloski et al.}
%
\institute{Jožef Stefan Institute, Ljubljana, Slovenia \and
Jožef Stefan International Postgraduate School \and 
Sapienza NLP Group, Sapienza University of Rome, Italy
\email{\{boshko.koloski;senja.pollak;blaz.skrlj\}@ijs.si\\navigli@diag.uniroma1.it}}

\maketitle        
%
\begin{abstract}
Large semantic knowledge bases are grounded in factual knowledge. However, recent approaches to dense text representations (i.e. embeddings) do not efficiently exploit these resources. Dense and robust representations of documents are essential for effectively solving downstream classification and retrieval tasks. This work demonstrates that injecting embedded information from knowledge bases can augment the performance of contemporary Large Language Model (LLM)-based representations for the task of text classification. Further, by considering automated machine learning (AutoML) with the fused representation space, we demonstrate it is possible to improve classification accuracy even if we use  low-dimensional projections of the original representation space obtained via efficient matrix factorization. This result shows that significantly faster classifiers can be achieved with minimal or no loss in predictive performance, as demonstrated using five strong LLM baselines on six diverse real-life datasets. The code is freely available at \url{https://github.com/bkolosk1/bablfusion.git}. 


\end{abstract}

\section{Introduction and Background}

Robust document representations are crucial for many NLP tasks \cite{muennighoff-etal-2023-mteb}. Early methods like bag-of-words were limited, relying on counting schemes and resulting in high-dimensional representations without capturing richer semantics. Techniques such as Latent Semantic Analysis (LSA) \cite{deerwester1990indexing} addressed this by projecting high-dimensional spaces into lower dimensions, providing more meaningful representations even in multilingual contexts \cite{koloski2020multilingual}. The representation learning paradigm \cite{replearn} popularized learning representations across modalities as an auxiliary task for training deep learning models. Le et al. \cite{le2014distributed} introduced Doc2Vec, which learns word or paragraph-level representations by corrupting text and predicting the missing parts using shallow neural networks. This technique remains key for obtaining document representations. Depending on how the corruption and learning are conducted, two main paradigms can be adopted: masked language modeling and causal language modeling. Devlin et al. \cite{devlin2018bert} demonstrated that randomly masking parts of the input (masked language modeling) and sequentially predicting them with the Transformer architecture \cite{vaswani2017attention} not only performs well but also learns contextual word embeddings. Conversely, Radford et al. \cite{radford_language_2019} approached document representation learning as a generative task, where a Transformer model \cite{vaswani2017attention} is fed part of the input and tasked with predicting the remainder. This training paradigm produces generative models and is currently the most popular approach towards LLMs \cite{zhao2023survey}. However, both paradigms focus on contextual word embeddings, which are insufficient for document-level representations. 
\par To leverage the expressiveness of deep models, Reimers et al. \cite{reimers-2019-sentence-bert} proposed using BERT-based embeddings as a foundation for learning document-level representations via Siamese networks. 
Similarly, LLM2Vec \cite{llm2vec} suggested representing documents by extracting the internal weights of large generative models, such as LLaMa3 \cite{llama3}. These embeddings can be efficiently obtained from a pre-trained model and serve as a strong competitor in a recently proposed massive text-embeddings benchmark (MTEB) \cite{muennighoff-etal-2023-mteb}.
Contrastive representation learning \cite{le2020contrastive} involves learning document representations by placing similar documents together and repelling dissimilar ones. 
Angle \cite{angle} was recently proposed, where models optimize representations based on the angle between their vectors in the latent space. However, high-dimensional representations can impair classifier performance due to the curse of dimensionality \cite{hughes}, increase memory footprint for storage and retrieval, and adapting these representations to specific corpora is laborious and expensive.
\par An alternating strand of work draws upon large semantic knowledge bases grounded in factual knowledge, such as Wikidata and BabelNet \cite{wikidata,navigli-ponzetto-2010-babelnet}. Koloski et al. \cite{neurocomputing} proposed a document representation approach that fused multiple transformer-based representations with a knowledge graph-grounded embedding. The method building on the knowledge enabled representations \cite{ostendorff2019enriching}, treated each n-gram tuple as a candidate entity, matched it to the knowledge graph, retrieved the embedding if present, and aggregated these embeddings into a single representation vector per document. These representations proved highly expressive for downstream tasks like multilingual semantic textual similarity assessment \cite{zosa2022embeddia}. However, the work did not explore document representations from generative and large language models \cite{llm2vec} or apply sophisticated entity linking and word sense disambiguation \cite{moro-etal-2014-entity}. Additionally, combining multiple representations is impractical for real applications due to the high-dimensional inputs negatively impacting classifier learning \cite{hughes}. One solution is to project high-dimensional inputs to a lower-dimensional space using dimensionality reduction methods like singular-value decomposition. Studies \cite{koloski2023latent,compersib} show that this procedure not only preserves the representations but also creates more representative spaces, further improving final-task performance. On the other side recent works show that contextual embeddings live on low-dimensional geometry \cite{hernandez-andreas-2021-low}. 
\par
\begin{figure}[H]
    \centering
    \includegraphics[width=\textwidth]{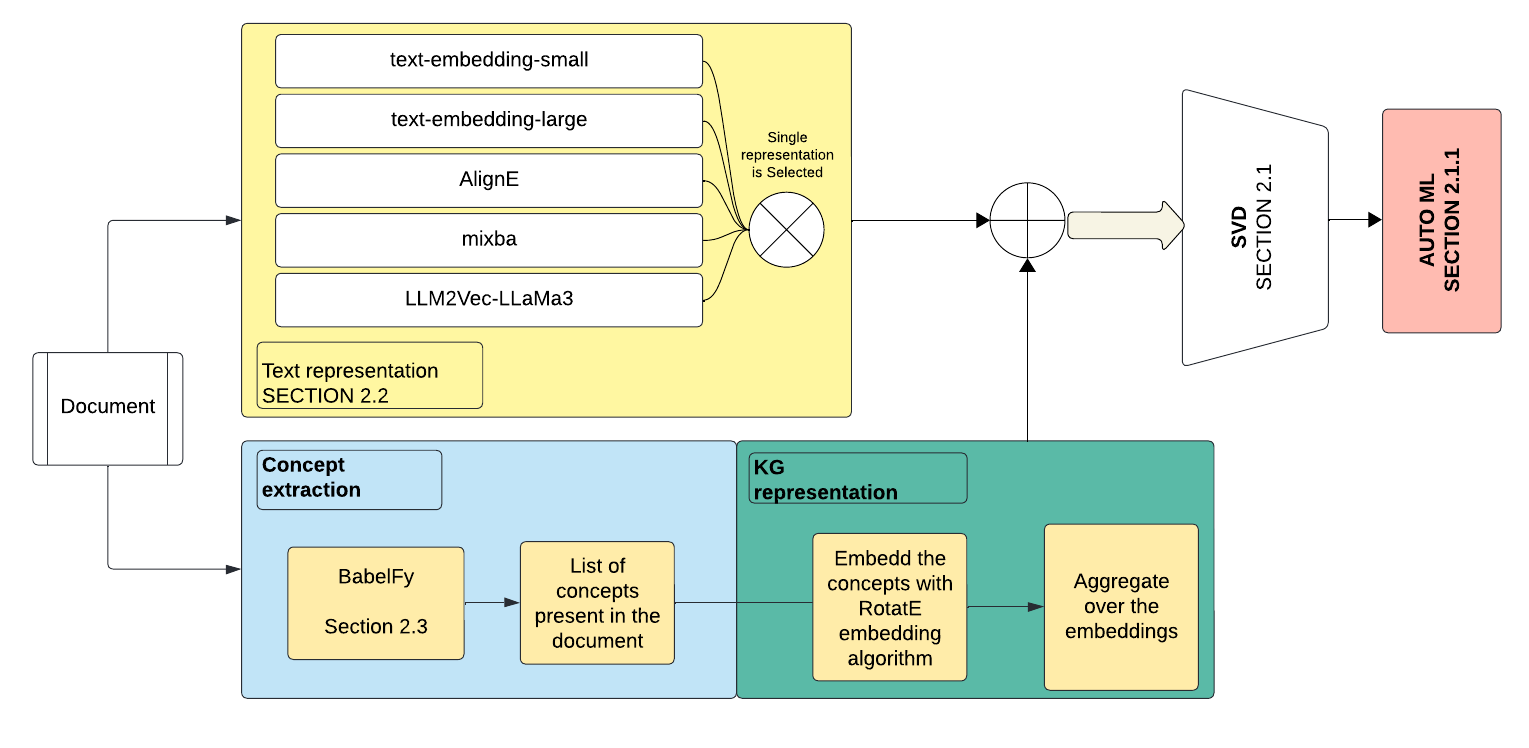}
    \caption{Schema of the proposed approach.}
    \label{fig:method}
\end{figure}

 A roadmap for unifying LLMs and knowledge bases was recently proposed \cite{Pan2023UnifyingLL} highlighting the potential of the symbiosis.  
Leveraging computational resources, evolutionary-based AutoML for learning document representations and models have achieved significant results \cite{autobot}.  
Motivated by these parallel approaches, we propose BabelFusion (see Figure~\ref{fig:method}), a novel approach towards document representation for classification where we leverage AutoML and low-dimensional projection of knowledge-informed representations, utilizing sophisticated entity linking \cite{moro-etal-2014-entity}. The novelty of this work can be summarized as follows: Firstly, to our knowledge, this is the first work that exploits the effect of injecting knowledge-based representations into LLM-based representations. Secondly, we demonstrate that, by projecting in low dimensions, one can learn robust and expressive representations, which, when combined with simple models, achieve competitive results in both full-shot and few-shot classification.
We present the methodology in Section \ref{sec:method}, followed by the experimental setting in Section \ref{sec:experim}. Section \ref{sec:results} presents the results which are followed by discussion in Section \ref{sec:disc}. 

\section{BabelFusion: Methodology}
\label{sec:method}
We proceed by discussing BabelFusion, the key contribution of this work.
Let $D=\{T,Y\}$ denote a dataset, where $T$ is a collection of textual documents and $Y$ is a collection of corresponding labels. Let $g$ be a representation learning function that maps the texts to a real-valued space of dimension $d$, such that $g(T) \mapsto X_{txt} \in \mathbf{R}^{d}$. 
Let $KG$ represent a knowledge graph, and let $k$ be a function that, for a given text ($t$), detects relevant entries in the knowledge graph and retrieves a list of vector representations of these detected entries, correlated with the text from the knowledge graph, such that $k(t) \mapsto \{e_{1}, e_{2}, \dots, e_{n}\}$, where each $e_{i} \in \mathbf{R}^{c}$. Having obtained a list of embeddings for a single document, we next average them to transform the collection of entity embeddings to a single vector in $\mathbf{R}^{c}$. This results in a representation of the texts as $X_{kg}$.

\subsection{Fusing text and knowledge graphs}

Given the text-based representation $X_{txt}$ and the corresponding knowledge graph representation $X_{kg}$, we aim to concatenate these representations to obtain richer text representations. This combined representation, denoted as $X_{\textrm{concat}}$, is obtained by concatenating the vectors from both sources:

$$ X_{\textrm{concat}} = [X_{\textrm{txt}} \mid X_{\textrm{kg}}] \in \textbf{R}^{d+c}$$

However, concatenating them directly and using them as concatenated results into higher $(d+c)$ dimensions which can degrade classifier performances as more dimensions can actually harm classifier performance due to the curse of dimensionality \cite{hughes,altman2018curse}. Thus, once we have $X_{\textrm{concat}}$, we apply Singular Value Decomposition (SVD) \cite{scikit-learn} to reduce its dimensionality and capture the most significant features. SVD decomposes $X_{\textrm{concat}}$ into three matrices: $U$, $\Sigma$, and $V$, such that:

$$ X_{\textrm{concat}} = U \Sigma V^T $$

Here, $U$ contains the left singular vectors, $\Sigma$ is a diagonal matrix with singular values, and $V$ contains the right singular vectors. To focus on the most relevant information, we perform a truncated SVD by selecting only the top $k$ singular values and their corresponding singular vectors. Mathematically, we truncate $\Sigma$ to $\Sigma_k$ by keeping only the $k$ highest singular values, and similarly truncate $U$ and $V$ to $U_k$ and $V_k$, respectively. By multiplying these truncated matrices, we obtain the final representation  $X_{\textrm{final}} = U_k \Sigma_k V^T_k \in \textbf{R}^{k} $. 

The truncation reduces the dimensionality of $X_{\textrm{concat}}$ while preserving the most important features \cite{compersib}.

\subsubsection{2.1.1 AutoML: Learning to classify}

To classify the documents into the $y$ labels, we fit a function $f$ over the representation $X_{\textrm{final}}$, such that $f(X_{\text{final}}) \mapsto y$. We usually learn by selecting over a family of functions with respect to some minimization of error. Specifically, we focus on applying the TPOT \cite{le2020scaling} library as an AutoML approach that leverages genetic algorithms to search the space of functions $f$ that minimize some error between the real labels (y) and the predicted labels $\hat{y}$, in our case the negative log loss defined as:
\[ \text{AUTOML}(\mathcal{L}(X_{\text{final}})), \quad \mathcal{L} = -\sum_{i=0}^{N}\sum_{j=0}^{|y|} y_j \log(\hat{y}_j) \]
where N is the number of documents and |y| is the number of classes. 
\vspace{-2em}
\subsection{Contemporary document representations}
\label{sec:text_reps}
Documents can be represented by both encoder-based and decoder-based large language models (LLMs). With this in mind, we explore several methods that map the documents into dense high-dimensional real space, $g(T) \mapsto X \in \mathbf{R}^{d}$. For decoder-based models, we use the recently proposed LLM2Vec paradigm \cite{llm2vec} to extract embeddings from the LlaMa3 model \cite{llama3}. For encoder-based models, we use the recently proposed Angle \cite{angle}, mxbai \cite{emb2024mxbai}, and two proprietary OpenAI embeddings (small and large) accessed via API\footnote{Accessed as of 14.7.2024}. More information can be found in Table \ref{tab:text_embs}, where we report the MTEB \cite{muennighoff-etal-2023-mteb} score.

\begin{table}[t]
\caption{Comparison of the used document representation models.}
\centering
    \label{tab:text_embs}
    \begin{tabular}{|l|l|l|c|}
        \hline
        Embedding &  Dimensions & Type & MTEB \cite{muennighoff-etal-2023-mteb} score (as of 14.7) \\
        \hline
        Angle \cite{angle} & 1024 & Encoder-only & 75.58 \\
        OpenAI-small & 1536 & Proprietary & 73.21 \\
        OpenAI-large &  3072 & Proprietary & 75.45   \\
        mxbai \cite{emb2024mxbai} &  1024 & Encoder-only & 75.64 \\
        LLM2Vec-LLaMa3 \cite{llm2vec} &  4096 & Decoder-only & 75.92 \\
        \hline
    \end{tabular}
\end{table}
\subsection{Knowledge representations}
\begin{figure}[htb!]
    \centering
    \includegraphics[width=0.8\textwidth]{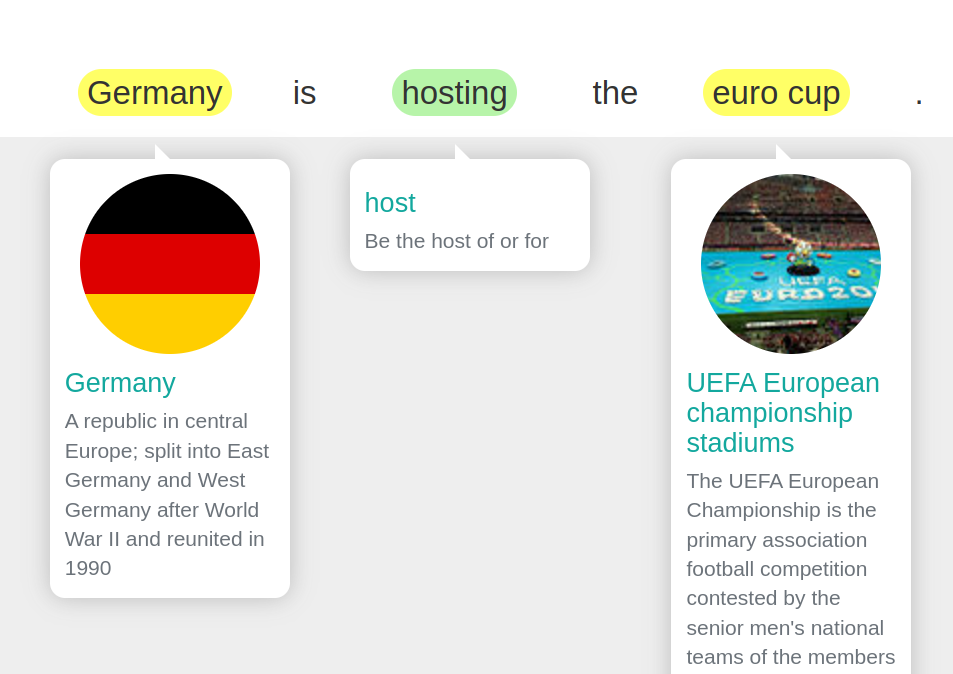}
    \caption{Babelfy disambiguation of the sentence \textit{Germany is hosting the euro cup.} The retrieved entities, are then matched to the WikiData5m sub-graph \cite{wikidata} and their respective embeddings are retrieved.}
    \label{fig:babelfy}
\end{figure}

In our experiments, we use the WikiData subgraph of BabelNet, which contains embedded nodes of the WikiData knowledge graph, using the RotatE \cite{sun2019rotate} method in a 512-dimensional real-valued space. We use GraphVite \cite{zhu2019graphvite}  to obtain the embeddings. First, we define the mapping function $k$, which produces a set of entities present in a knowledge graph. We employ Babelfy \cite{moro-etal-2014-entity}, an algorithm that operates on the following principle: Given a lexicalized semantic network (BabelNet) and an input text, Babelfy identifies all linkable fragments. It then performs a graph-based semantic interpretation, constructing a graph where nodes represent candidate meanings and edges denote semantic relationships. The algorithm extracts the densest subgraph as the best candidate meaning for each fragment. The resulting output is a list of these candidate meanings, providing a coherent semantic representation of the input text an example of one such mapping is presented in Figure \ref{fig:babelfy}.

\section{Experimental setup}
\label{sec:experim}
In this section we present the experimental setup. We present the dataset s for evaluation in Section \ref{sec:dataset}, followed by the evaluation setup in Section \ref{sec:eval}.
\subsection{Datasets}
\label{sec:dataset}
We aim to evaluate the proposed method in two distinct classification domains: sentiment analysis and news genre classification. For sentiment analysis, we utilize the standard Amazon Reviews for sentiment Analysis dataset, which includes reviews from three different subforums (Books, DVD, and Music), as well as a hate speech classification dataset consisting of short social media posts categorized into hate speech and non-hate speech \cite{ranasinghe-zampieri-2020-multilingual}. For news genre classification, we employ the MLDoc \cite{mldoc} dataset, which categorizes news into four genres, and the recently proposed XGENRE \cite{kuzman-etal-2022-ginco} dataset. 
In summary, we use six classification datasets: four binary classification datasets for sentiment analysis and two multi-class datasets for news genre classification. We use the original train-test splits per dataset. Table \ref{tab:data} presents more in-depth dataset statistics. 

\begin{table}[h!]
    \centering
    \caption{Datasets considered (with statistics).}
    \begin{tabular}{c|c|c|r|r|r}
        Dataset & Domain & Labels & Train documents & Test documents & Avg word count    \\ 
        \midrule
        Books & sentiment & 2 &  2000 & 2000 & 155.80\\
        Dvd & sentiment & 2 & 2000 & 2000 & 161.29 \\
        Music & sentiment & 2 &2000 & 2000  & 130.12\\
        Hate speech & sentiment & 2 & 13240 & 860  & 22.85 \\
        MLDoc & news & 4 & 11000 & 4000 & 235.15 \\
        XGENRE & news & 9 & 1650 & 272 & 1256.92 \\        
    \end{tabular}
    \label{tab:data}
\end{table}
\subsection{Evaluation setup}
\label{sec:eval}
We aim to evaluate the performance potential of knowledge-induced, low-dimensional representations for document classification.
\vspace{-1em}
\subsubsection{Baselines}
Our objective is to enhance document representation quality. The baseline method involves training a linear classifier with ridge regression penalization on text representations (see Section \ref{sec:text_reps}). The choice for this baseline is that related work has shown that these representations are powerful on their own and the penalization of ridge regression is sufficient to obtain competitive results.
\textbf{End-to-end}
We assess the performance of fused representations versus text-only representations with full data availability. 
\textbf{Few-shot}
We evaluate the performance with limited training data using stratified subsampling at 1\%, 2\%, 5\%, 10\%, 20\%, 50\%, and 100\% of the available data.
\textbf{Learning in low dimensions}
Projecting into lower dimensions can both enhance and deteriorate representations \cite{koloski2023latent}. We explore the effects on our representations compared to text-only representations by projecting them into 2, 4, 8, 16, 32, 64, 128, and 512 dimensions.
We address the following research questions: 
\begin{itemize}
    \item Q1. Do knowledge-enriched document representations consistently outperform text-based representations?
    \item Q2. Is learning in low dimensions (projected representations) as expressive as learning in high dimensions?
    \item Q3. Which family of models benefits more from knowledge-based enhancement, encoder- or decoder- only? 
    \item Q4. Can we improve proprietary, state-of-the-art LLM-based embeddings with the introduction of external (KG-based) knowledge?   
\end{itemize}

We use the HuggingFace \cite{wolf-etal-2020-transformers} library to obtain the document representations in conjunction with the sentence-transformers library \cite{reimers-2019-sentence-bert} library. For the AUTOML search, per each dataset and text representation (see Section \ref{sec:text_reps}), we allow up to 1-hour run-time, $256GB$ of RAM and a max of $16$ cores. We search for up to $100$ generations, with $100$ samples per population. For fitting the AutoML learner we perform 5-fold cross-validation. We use Logistic Regression implementation in  \cite{scikit-learn} for the baseline ridge regression. For obtaining OpenAI embeddings we use their API\footnote{\url{https://platform.openai.com/docs/guides/embeddings/use-cases}}. For the remaining embeddings we use the default settings and obtain them thorough Huggingface \cite{wolf-etal-2020-transformers}.

\section{Results}
\label{sec:results}

We proceed by discussing results of experimental evaluation outlined in Section~\ref{sec:experim}.

\subsection{End-to-end classification}

We present the results of the best performing BabelFusion approach compared to the baseline Ridge classifier over the text in high dimensions in Table \ref{tab:vertical_split}. Our proposed method outperformed the baseline on average by $0.52\%$, with the difference being statistically significant as per the  Wilcoxon Signed-Rank Test (statistic = 98.0, p-value = 0.01).

\begin{table}[H]
    \centering
    \caption{Accuracy of Document Representations Across Datasets. \underline{Underlined} entries indicate the model that outperformed others in the given setting, while \textbf{bolded} entries highlight the best overall model.}

    \resizebox{\textwidth}{!}{\begin{tabular}{l||cc|cc|cc|cc|cc|cc}
\toprule
Dataset  & \multicolumn{2}{c}{Books} & \multicolumn{2}{c}{DVD} & \multicolumn{2}{c}{Music} & \multicolumn{2}{c}{Hate speech} & \multicolumn{2}{c}{MLDoc} & \multicolumn{2}{c}{XGENRE} \\
Representation & baseline & ours & baseline & ours & baseline & ours & baseline & ours & baseline & ours & baseline & ours  \\
\midrule
Angle & 93.85 & \underline{95.40} & 94.15 & \underline{94.95} & 91.65 & \underline{94.25} & 79.06 & \underline{81.62} & 95.42 & \underline{95.90} & 53.67 & \underline{\textbf{59.19}} \\
LLaMa3 & 92.45 & \underline{93.65} & \underline{92.05} & 92.00 & 91.65 & \underline{92.95} & 76.74 & \underline{79.18} & \underline{96.52} & 96.15 & \underline{57.72} & 56.98 \\
OpenAI-large & 93.95 & \underline{\textbf{96.05}} & 94.15 & \underline{95.15} & 93.75 & \underline{\textbf{95.25}} & \underline{\textbf{83.72}} & 75.11 & 96.37 & \underline{\textbf{97.15}} & 54.77 & \underline{55.14} \\
OpenAI-small & 94.00 & \underline{94.15} & \underline{94.15} & 93.90 & \underline{93.75} & 93.70 & \underline{\textbf{83.72}} & 76.97 & 96.37 & \underline{96.87} & \underline{54.77} & 53.30 \\
mxbai & 94.00 & \underline{95.75} & 94.10 & \underline{\textbf{95.35}} & 91.90 & \underline{94.00} & 79.53 & \underline{81.04} & 95.72 & \underline{96.30} & 55.51 & \underline{57.35} \\ \bottomrule
average & $93.65 _{ 0.67}$  & $\textbf{ 95.00} _{ 1.04}$ & $93.72 _{ 0.93}$ & $\textbf{94.27} _{ 1.38}$ & $96.08 _{ 1.10}$  &  $\textbf{96.47}  _{ 0.83}$ &  $\textbf{80.55} _{3.07}$ &  $78.78 _{ 2.74}$ & $92.54 _{ 0.48}$  & $\textbf{94.03} _{ 0.51}$ & $55.28 _{ 1.50}$ & $\textbf{56.39}_{  2.24}$ \\
\end{tabular}}
    \label{tab:vertical_split}
\end{table}

Next, we assess the gains and their statistical significance across representations via the t-Test statistics. We notice the biggest gain for the Angle embedding 2.25$\pm$1.85 percentage points, statistically significant (paired t-Test statistics=-3.02, p-value=0.03), followed by mxbai (1.5$\pm$0.54) points statistically significant (paired t-Test statistics=-6.85, p-value=0.01)) and LLM2Vec-LLaMa3 (0.63$\pm$1.22). On average, we notice a minor decrease for the two OpenAI variants, small (-1.31$\pm$2.75) and large (-0.48$\pm$4.03). This discrepancy originates from the differences in the hate speech datasets, where we may have destroyed the power of the initial embeddings due to the nature of the informal speech of online debates. 
\par We compare the performance of our method across the two domains: sentiment and news. First, we perform the Shapiro-Wilkinson test to assess if the differences are normally distributed, which in our case are not; more precise statistics for domain sentiment (W-statistic=0.81, p-value=0.02) and news (W-statistic=0.64, p-value<0.01). Following this, we employ the Mann-Whitney U test, which shows that the difference across the two domains is not statistically significant (statistics=71.0, p-value=0.21), suggesting that our method is applicable across domains.


\subsection{Impact of dimensionality reduction}
\begin{figure}[htb]
    \centering
    \includegraphics[width=\textwidth]{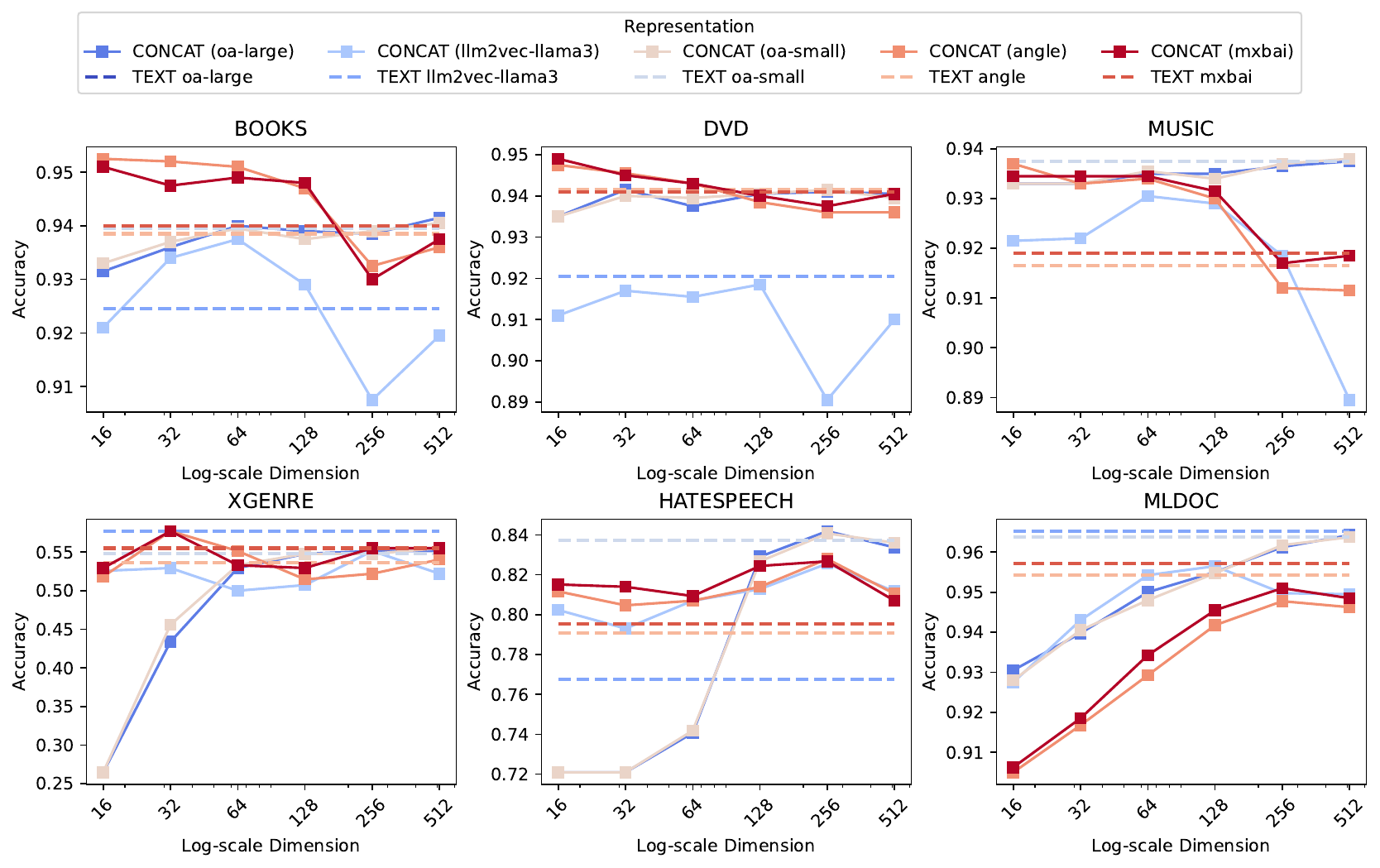}
    \caption{Projecting at different dimensions. The x-axis is log-scaled for better portrial of results.}
    \label{fig:low-dim}
\end{figure}

Next, we analyse the impact of the projected dimension $c$ on performance. We show the results in Figure \ref{fig:low-dim}. Learning in lower dimensions for the hate speech, genre, and MLDoc datasets shows lower results for all embeddings, as expected. Interestingly, for the Amazon datasets, where some embeddings (mxbai and oa-small) outperform the full-text-based representation baselines, we can learn even if we project in two dimensions. We also find that on all datasets we can outperform baselines for all methods on different dimensions, even on the more difficult datasets hate speech and XGENRE.
We examined the correlation between the dimension and the score across all embeddings and found no statistical correlation, implying that the dimensionality of the projection is crucial and should be evaluated for each dataset and problem. We then analysed the correlation between each dataset's dimension and score. We find that it is significant for the Hate speech dataset (correlation=0.62, p-value<0.01,
 CI-95=[0.40, 0.78]), the XGENRE dataset (correlation=0.52, p-value<0.01, CI-95=[-0.26,0.33]) and MLDoc (correlation=0.47, p-value=0.01, CI-95=[0.20, 0.67]).

\begin{figure}[htb]
    \centering
        \includegraphics[width=0.7\textwidth]{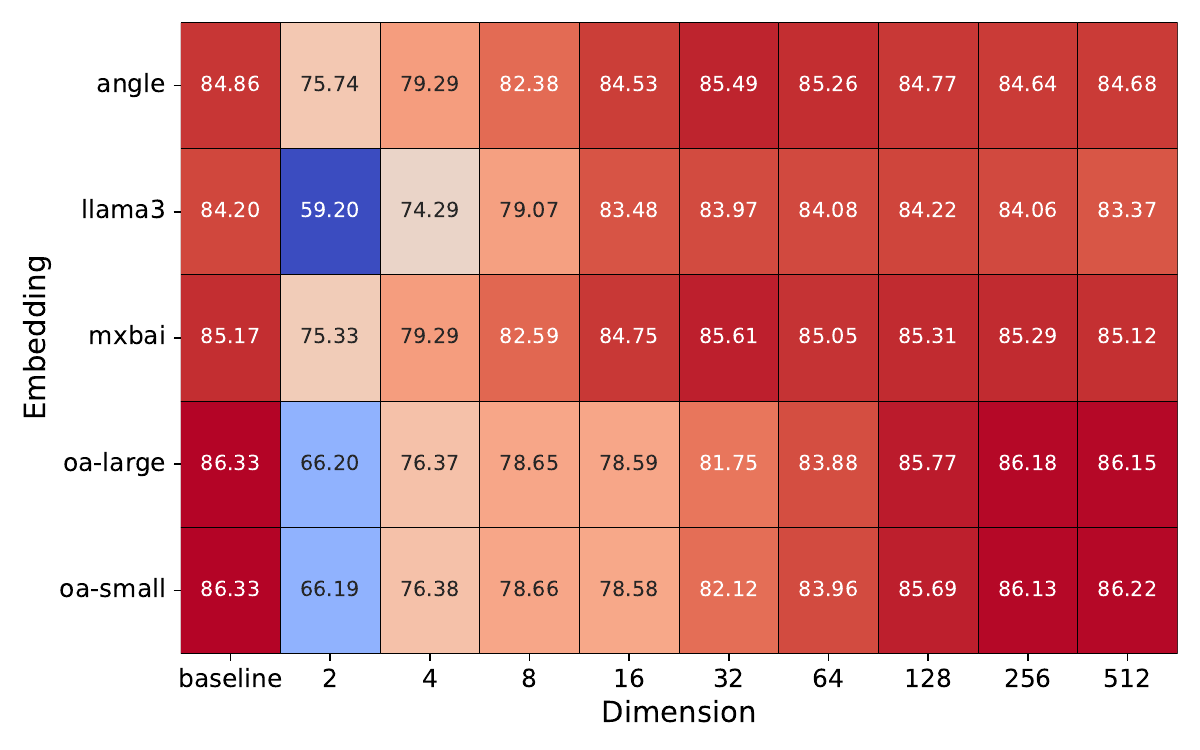}
        \caption{Aggregated results for each embedding across dimensions.}
        \label{fig:plot1}
\end{figure}

\begin{figure}[htb]
    \centering
    \centering
    \includegraphics[width=0.7\textwidth]{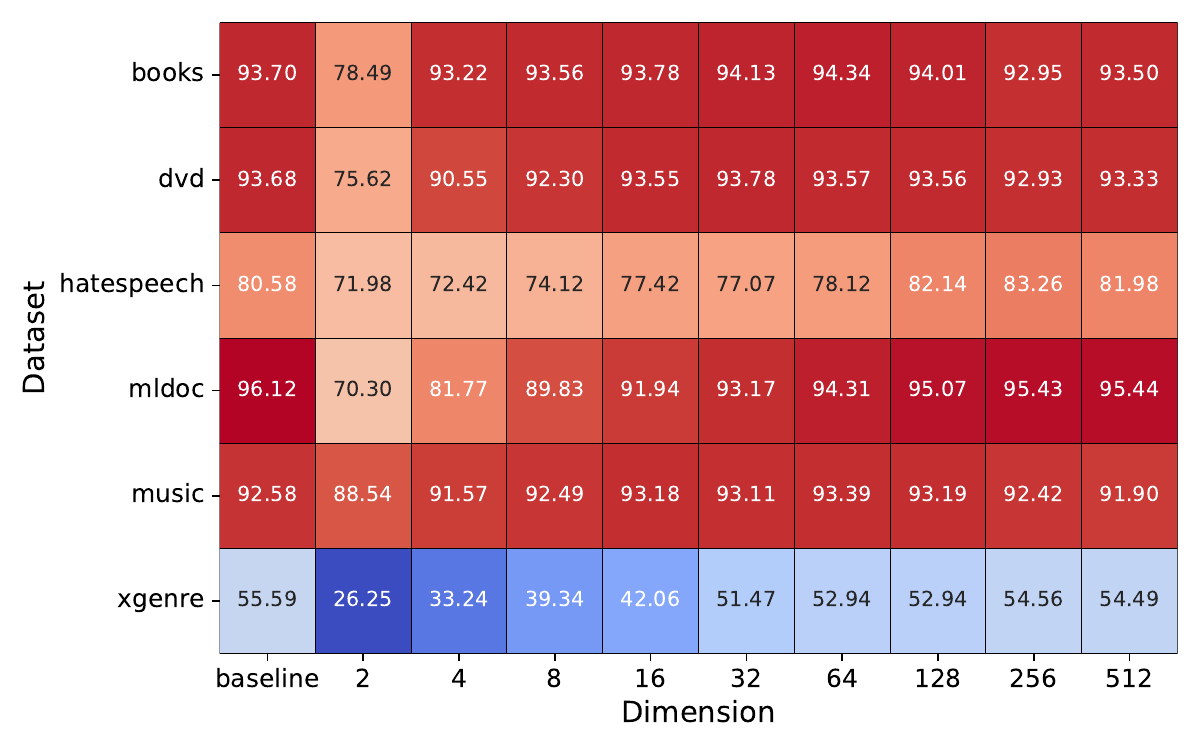} 
    \caption{Aggregated results for each dataset across dimensions.}
    \label{fig:plot2}
\end{figure}

Next, we aggregate the results across dimensions for Embeddings (Figure\ref{fig:plot1}) and for Datasets (Figure \ref{fig:plot2}) and compare them to the outcomes when learning occurs in the joint space without any projection (the left-most column in the heatmaps, labeled as 'baseline').

We see that across embeddings, we can learn more robust spaces by injection of embedded entities and projection to low dimensions, meaning that we cannot only learn in low-dimensional space but also obtain better results. This follows the related work by \v{S}krlj et al. \cite{compersib}, where it was shown that compressing the space lowers the memory footprint and can improve the end performance. Across datasets, we notice that learning from low dimensions is also, on average, better than learning from high, as learning from low dimensions improves the results. 


\subsection{Few-shot learning}
\begin{figure}[htb]
    \centering
    \includegraphics[width=0.9\textwidth]{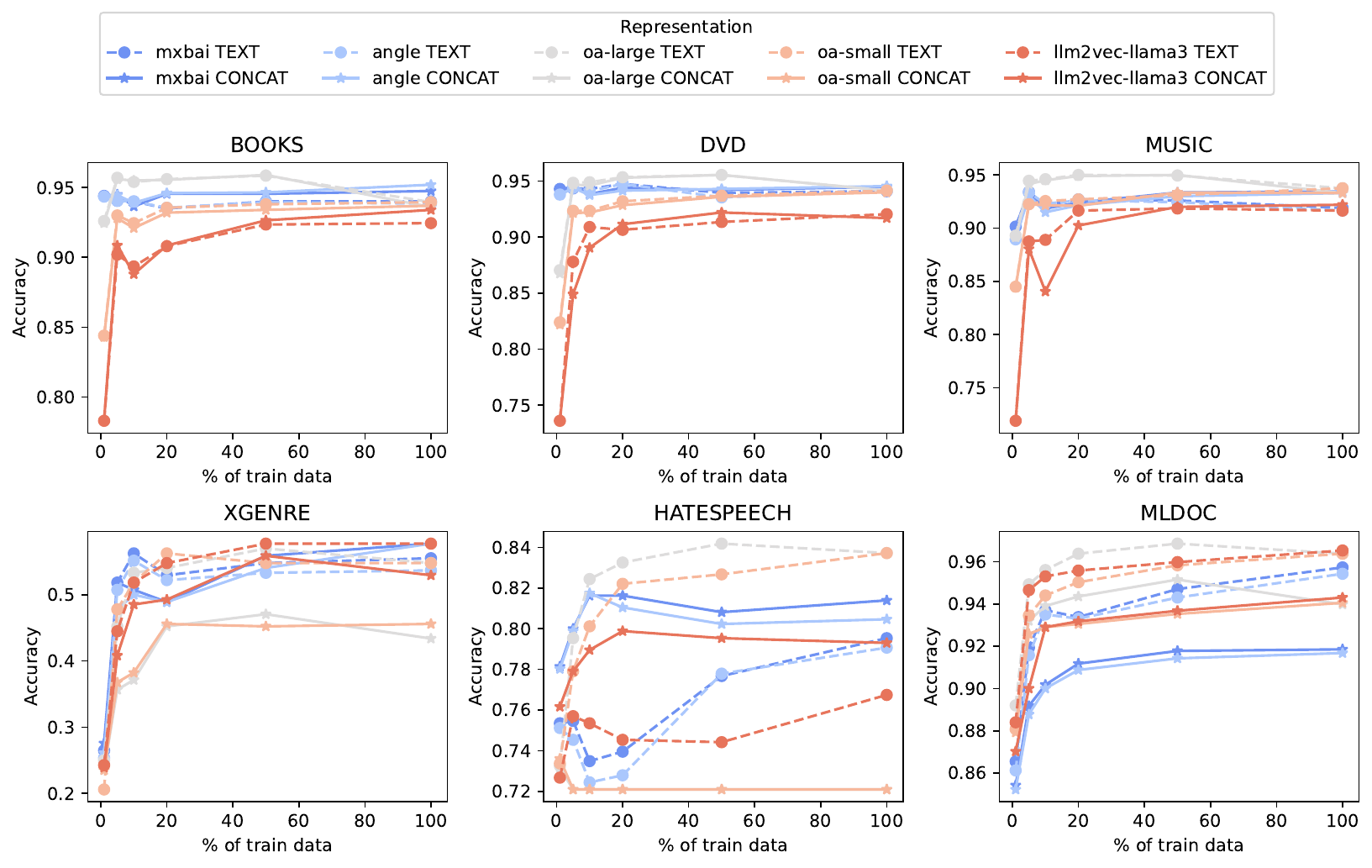}
    \caption{Few-shot classification results.}
    \label{fig:clf-fewshot}
\end{figure}

In Figure \ref{fig:clf-fewshot}, we show the method's performance on fractions of data. The results indicate that the method performs on par compared to text-only baselines on the same fraction of data. For some datasets (DVD, music and books), the method achieved better results with a smaller sample (compared to the full-shot approach). Recent works \cite{gao2021making,lin2023chatqa} have shown that this can be the case for LLMs when applied to downstream tasks, and as many of our methods are based on LLMs, we believe this might be the case as well. We see a more considerable discrepancy with the results for the XGENRE, hate speech and MLDoc datasets, probably because these documents come from more versatile distribution, making the problem harder, and more examples alleviate that problem.

\subsection{Qualitative results}

\subsubsection{Statistics of the retrieved KG entries} Next, we explore the statistics of the retrieved entries in WikiData as matched by Babelfy \cite{navigli-ponzetto-2010-babelnet}, as shown in Table \ref{tab:stats_bbnet}. For the datasets derived from news corpora, we observe a higher extraction of concepts. We attribute this to the standardized language typically used in news writing, in contrast to the non-standard language, slurs, and typos prevalent in social media posts, as noted in the seminal paper on the hate speech dataset \cite{ranasinghe-zampieri-2020-multilingual}. We also note that the hate speech dataset, containing the shortest documents, had approximately 22\% of its documents without any matched entity against the knowledge graph. The high number of detected entries in BabelNet for the XGENRE and MLDoc datasets reflects the nature of the data, as news articles tend to be longer on average. The results indicate that the nature and length of the text significantly influence the number of matched entities. 
\begin{table}[H]
    \centering
    \caption{Statistics of retrieved entries from WikiData with Babelfy per dataset.}
    \begin{tabular}{l|r|c|c|c|c}
        Dataset     & Docs without & Mean entries & Max entries & Min entries & Median entries \\
        \toprule
        Music       & 2\%  & 18.03 $\pm$ 19.27 & 189 & 0 & 12 \\
        DVD         & 3\%  & 21.34 $\pm$ 25.08 & 272 & 0 & 13 \\
        Books       & 3\%  & 19.00 $\pm$ 22.41 & 203 & 0 & 12 \\
        Hate speech  & 22\%  & 2.83 $\pm$ 2.63   & 22  & 0 & 2 \\
        XGENRE      & 0\%  & 61.90 $\pm$ 48.08 & 281 & 3 & 49 \\
        MLDoc       & 0\%  & 48.04 $\pm$ 37.01 & 440 & 2 & 37 \\
    \end{tabular}
    \label{tab:stats_bbnet}
\end{table}
 \vspace{-1.8em}

\subsubsection{Visualization of the embeddings}

\begin{figure}[H]
    \centering
    \includegraphics[width=\textwidth]{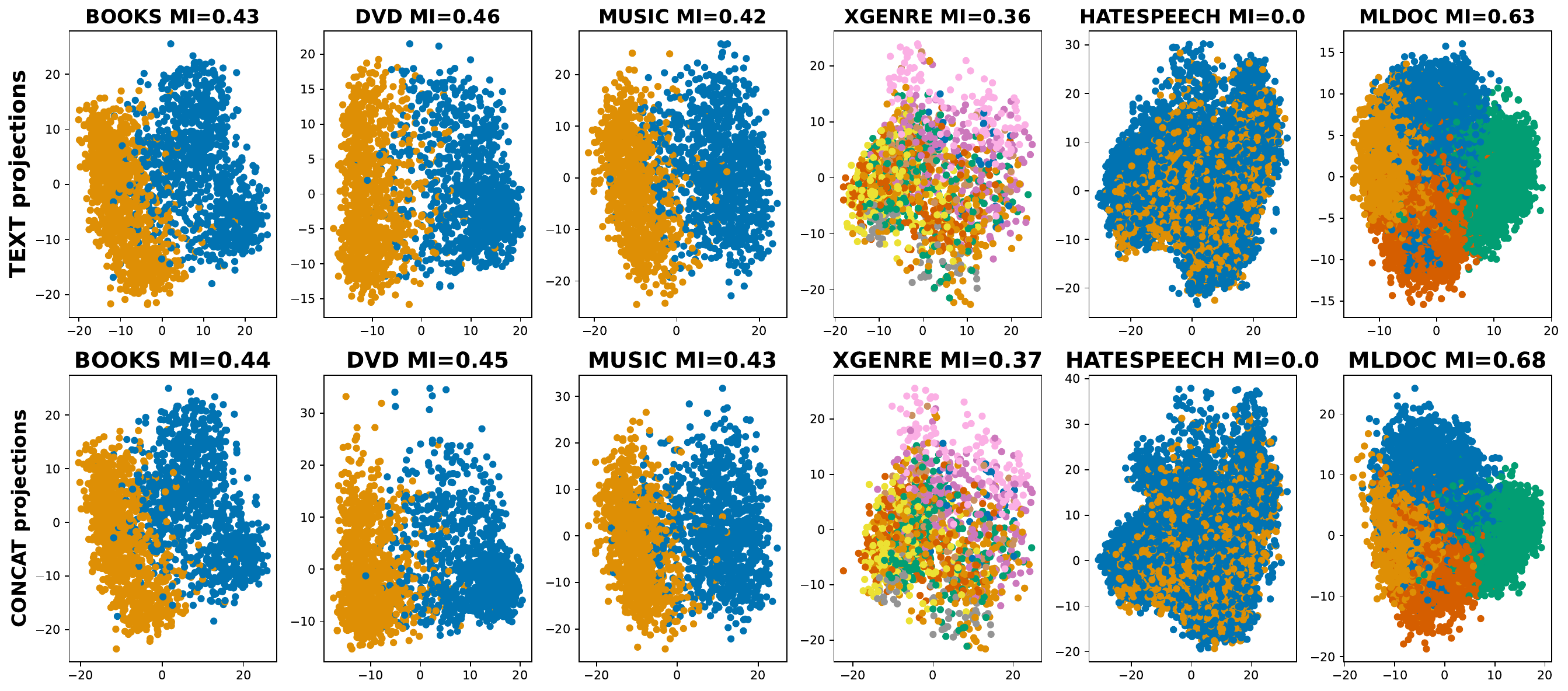}
    \caption{2D visualization of the text embeddings compared to the concatenated embeddings. Color indicates the class label. Best viewed on screen. }
    \label{fig:qualitative}
\end{figure}

We visualize the embeddings for all datasets in Figure \ref{fig:qualitative}. In the top row, we represent the text-only representation $X_{txt}$ reduced to two dimensions with SVD, while in the second row, we present the concatenated embeddings $X_{\textrm{concat}}$ reduced to two dimensions with SVD. In addition, we perform K-means clustering (K = number of classes) over the projected text and image embeddings in two dimensions, and report the achieved scores in the titles. What we observe is that the projection of the enriched embeddings performs on par or better, when considering Normalized Mutual Information, meaning that this space also qualitatively enables better separation. This finding suggests that one can use enriched embeddings for all kinds of tasks that require this property, such as clustering and topic modeling.

\section{Conclusions and Further Work}
\label{sec:disc}
In this work, we propose new knowledge-enriched, LLM-based low-dimensional document embeddings. The results suggest that fusing modalities in low dimensions not only preserves space but also enables efficient representations that surpass even proprietary embeddings. This is in line to an extent with the work of \cite{hernandez-andreas-2021-low} where it was shown that contextual embeddings can be approximated by low-dimensional geometry. We advance the related line of work by introducing sophisticated named entity linking and AutoML while leveraging representations extracted from LLMs. Our findings demonstrate that these embeddings perform well across different datasets and domains, showing promising potential for future applications. The results indicate that,  even in unsupervised applications, such as clustering, these lightweight embeddings might provide robust document representations. Furthermore, by applying AutoML, we show that learning in low dimensions is feasible and competitive with high-dimensional embeddings. 
\par The implications of these results are that: \textbf{A1.} KG-enriched representations can outperform text representations, including both encoder and decoder-based models (\textbf{A3}), and that learning on these representations in low dimensions is feasible (\textbf{A2}),  even for proprietary document representations (\textbf{A4}).
\par In this study, we utilized only a portion of the BabelNet graph and the WikiData5m subgraph. In the future, we aim to include the entire graph, improve the fusion process with advanced disambiguation methods, and explore injection of external knowledge at the token level to create a synergy between LLMs and KGs, as suggested by Pan et al.  \cite{Pan2023UnifyingLL}. Finally, we plan to explore if recursive compression of our proposed representations provides better down-stream results.


\vspace{-1em}

\section*{Acknowledgments}
\vspace{-1em}
The authors acknowledge financial support from the Slovenian Research and Innovation Agency through research core funding (No. P2-0103) and projects No. J4-4555, J5-3102, L2-50070, and PR-12394, and from CREATIVE project (CRoss-modal understanding and gEnerATIon of Visual and tExtual content) funded by the MUR Progetti di Ricerca di Rilevante Interesse Nazionale programme (PRIN 2020). 
\bibliographystyle{splncs04}
\bibliography{bibliograph}
\end{document}